\documentclass{myifcolog}

\usepackage{amsmath,amssymb,amsthm}
\usepackage{braket}
\usepackage{mathrsfs}
\usepackage{stmaryrd}
\usepackage{subcaption}

\usepackage{natbib}

\usepackage{tikz}
\usepackage{lipsum}

\newcommand{\CPM}[1]{\mathbf{CPM}(#1)}
\newcommand{\FVect}{\mathbf{FVect}}
\newcommand{\ConvexRel}{\mathbf{ConvexRel}}

\newcommand{\CPe}[1]{\mathcal{CP}_1(#1)}
\newcommand{\semantics}[1]{\llbracket #1 \rrbracket}
\newcommand{\lang}[1]{\ensuremath{\textit{#1}}}

\newcommand{\loewner}{\sqsubseteq}
\newcommand{\I}{\mathbb{I}}

\titlethanks{}

\title{Towards logical negation in compositional distributional semantics}
\titlerunning{Towards negation in DisCoCat}
\addauthor[m.a.f.lewis@uva.nl]
  {Martha Lewis}
  {ILLC, University of Amsterdam}
\authorrunning{Martha Lewis}

\usepackage{tikz}
\usepackage{pgfplots}
\usetikzlibrary{trees}
\usetikzlibrary{topaths}
\usetikzlibrary{decorations.pathmorphing}
\usetikzlibrary{decorations.markings}
\usetikzlibrary{matrix,backgrounds,folding}
\usetikzlibrary{chains,scopes,positioning,fit}
\usetikzlibrary{arrows,shadows}
\usetikzlibrary{calc} 
\usetikzlibrary{chains}
\usetikzlibrary{shapes,shapes.geometric,shapes.misc}

\pgfdeclarelayer{edgelayer}
\pgfdeclarelayer{nodelayer}
\pgfsetlayers{background,edgelayer,nodelayer,main}

\tikzstyle{dot}=[circle,fill=black,draw=black]
\tikzstyle{none}=[inner sep=0pt]
\tikzstyle{every loop}=[]

\tikzstyle{(null)}=[]
\tikzstyle{plain}=[]

\tikzstyle{blank}=[inner sep=0pt]
\tikzstyle{box}=[rectangle, minimum size = 0.5cm,fill=white,draw=black]
\tikzstyle{small_node}=[inner sep=0pt, minimum size=0.2cm,circle,fill=white,draw=black]
\tikzstyle{square}=[thick, minimum size=0.3cm,rectangle,fill=white,draw=black]
\tikzstyle{small_circle}=[inner sep=0pt, minimum size=0.25cm,circle,fill=white,draw=black]
\tikzstyle{big_circle}=[inner sep=0pt, minimum size=2cm,circle,fill=none,draw=black]
\tikzstyle{big_rectangle}=[inner sep=0pt, thick, rounded corners, minimum height=2cm, minimum width=2.5cm,rectangle,fill=none,draw=black]
\tikzstyle{circle_node}=[inner sep=0pt, minimum size=0.5cm,circle,fill=white,draw=black]
\tikzstyle{medium_circle}=[inner sep=0pt, minimum size=1.5cm,circle,fill=none,draw=black]
\tikzstyle{vbig_rectangle}=[inner sep=0pt, thick, rounded corners, minimum height=5.5cm, minimum width=5.5cm,rectangle,fill=none,draw=black]

\tikzstyle{downtri}=[regular polygon,regular polygon sides=3,shape border rotate=180,fill=white,draw=black]
\tikzstyle{uptri}=[regular polygon,regular polygon sides=3,shape border rotate=0,fill=white,draw=black]

\tikzstyle{morph}=[->,draw=black,line width=0.600]
\tikzstyle{arrow}=[-,draw=black,postaction={decorate},decoration={markings,mark=at position .5 with {\arrow{>}}},line width=2.000]
\tikzstyle{tick}=[-,draw=black,postaction={decorate},decoration={markings,mark=at position .5 with {\draw (0,-0.1) -- (0,0.1);}},line width=2.000]

\begin{document}
\maketitle
\begin{abstract}
The categorical compositional distributional model of meaning gives the composition of words into phrases and sentences pride of place. However, it has so far lacked a model of logical negation. This paper gives some steps towards providing this operator, modelling it as a version of projection onto the subspace orthogonal to a word. We give a small demonstration of the operators performance in a sentence entailment task.
  
\end{abstract}

\section{Introduction}
Compositional models of meaning aim to represent the meaning of phrases and sentences by combining representations of the words in the sentence according to some rule. Compositional distributional models, such as described in \cite{baroni2010, coecke2010, paperno2014} combine the compositional approach with vector-based models of word meaning. In these models, nouns are represented as vectors, and function words, such as verbs and adjectives, are modelled as linear maps. In this paper, we use the \emph{categorical compositional distributional} (\emph{DisCoCat}) model introduced in \cite{coecke2010}. This model formalises the compositional approach to language using category theory, setting up a functorial mapping between the grammar of the language on the one hand, and the structures used to represent lexical meaning on the other. In modelling the meaning of words and sentences, a distinction can be made between words with lexical content, and words that can arguably be modelled as an operation on the structure of the sentence. For example, in \cite{sadrzadeh2013}, relative pronouns are modelled as routing information around a sentence using the structure of a Frobenius algebra. In \cite{kartsaklis2016co}, conjunctions are modelled using Frobenius algebras. 

In the current paper we model negation as an operation on words. In \cite{coecke2010}, negation is modelled as a linear map on a two-dimensional sentence space which sends each basis vector to the subspace orthogonal to it. This idea of modelling negation as projection to the orthogonal subspace was used in \cite{widdows2003}, but at the vector level is somewhat unsatisfactory since a word and its negation are then of two different kinds. Furthermore within DisCoCat, words should be modelled as linear maps, which projection onto an orthogonal subspace doesn't satisfy.

Within the categorical compositional framework, we can be flexible about how to represent word meanings. In \cite{coecke2010} the category $\FVect$ of vector spaces and linear maps was used, meaning that nouns and sentences are modelled as vectors, and functional words such as verbs and adjectives are modelled as linear maps. In \cite{bolt2017} the category $\ConvexRel$ was used, enabling the representation of nouns and sentences as convex sets and function words as convex relations. In this paper we will use the category $\CPM{\FVect}$ which models nouns and sentences as positive operators, and function words as completely positive maps. This approach to meaning was developed in \cite{balkir2016, bankova2019} and implemented in \cite{lewisfthc, lewis2019}. We model negation as an operation related to projection onto the orthogonal subspace. We discuss how negation interacts with composition, and we provide a small corpus-based implementation to illustrate the ideas.

\subsection{Related work} As mentioned, the idea of negation as projection onto the orthogonal subspace has been implemented in \cite{widdows2003} and discussed in \cite{coecke2010}. However, it has also been argued that negation should not be viewed in this way: rather, that the negation of a word should be fairly similar to the original. for example, \cite{hermann2013} argue that `not red' is still a colour, and provide a model where the vector is divided into domains and only part of the vector is inverted. Similarly, \cite{rimellneg} view negation as antonymy, and provide a model of negation in which an encoder is trained to produce the antonym of a given adjective. Continuing the discussion of the distinction between conversational negation and logical negation, \cite{kruszewski2016} provide an in depth analysis of the ways in which people use negation in conversation.

The kind of negation that we discuss in this paper is more akin to logical negation, and this will be exemplified by its interaction with entailment between sentences.

\section{Background}
\subsection{Categorical compositional approaches to meaning}
The categorical compositional model of meaning uses the framework of category theory to set up a mapping between the grammar of a language and the structures used to represent the manings of individual words. A formalization of grammar is chosen, and represented as a category, called the \emph{grammar category}. A choice is made about the type of meaning representation, which again is formalized as a category, called the \emph{meaning category}. The meaning category and the grammar category are chosen to have the same abstract structure. Type reductions in the grammar category are then functorially mapped to operations in the semantics category. In this paper, the grammar category and the meaning category are both \emph{compact closed}. For details of what this means within the context of linguistics, see \cite{coecke2010} or \cite{preller2011}. A gentle presentation is also given in \cite{bolt2017}.

\paragraph{Pregroup grammar} In this paper, we will use pregroup grammar, although the formalism is flexible about what can be used, and other choices are given in, for example, \cite{lambekvslambek, maillardccg, muskens2016}. A pregroup is a partially ordered monoid $(X, \cdot, 1, \leq)$ where each $x \in X$ has a left and a right adjoint $(-)^l$, $(-)^r$ such that: 
  \begin{equation}
    \epsilon^r_x:x\cdot x^r \leq 1, \qquad \epsilon^l_x:x^l\cdot x\leq 1 \qquad \eta^r_x:1 \leq x^r\cdot x, \qquad
    \eta^l_x:1\leq x\cdot x^l
  \end{equation} 
A pregroup grammar is the pregroup freely generated over a set of chosen types. We consider the set containing $n$ for noun and $s$ for sentence. Complex types are built up by concatenation of types, and we often leave out the dot so that $xy = x\cdot y$. If $x\leq y$ we say that $x$ \emph{reduces to} $y$.

A string of types $t_1, ... t_n$ is \emph{grammatical} if it reduces, via the morphisms above, to the sentence type $s$. For example, typing $\lang{clowns}$ as $n$, $\lang{tell}$ as $n^r s n^l$ and $\lang{the truth}$ as $n$, the sentence $\lang{Clowns tell the truth}$ has type $n (n^r s n^l) n$ and is shown to be grammatical as follows:
\begin{align}
(\epsilon^r \ 1 \ \epsilon^l) n (n^r s n^l) n &\leq (\epsilon^r \ 1)(n\ n^r s \ 1) \\
&\leq 1\ s \ 1 = s
\end{align}

The above reduction can be represented graphically as follows:
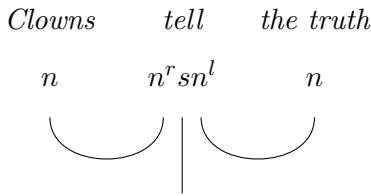
\begin{center}
\begin{tikzpicture}
	\begin{pgfonlayer}{nodelayer}
		\node [style=none] (0) at (0, -0) {};
		\node [style=none] (1) at (-0.25, -0) {};
		\node [style=none] (2) at (0.25, -0) {};
		\node [style=none] (3) at (1.75, -0) {};
		\node [style=none] (4) at (-1.75, -0) {};
		\node [style=none, anchor=mid] (5) at (-1.75, 0.5) {$n$};
		\node [style=none, anchor=mid] (6) at (0, 0.5) {$n^r s n^l $};
		\node [style=none, anchor=mid] (7) at (1.75, 0.5) {$n$};
		\node [style=none] (8) at (0, -1) {};
		\node [style=none, anchor=mid] (9) at (-1.75, 1.25) {\lang{Clowns}};
		\node [style=none, anchor=mid] (10) at (0, 1.25) {\lang{tell}};
		\node [style=none, anchor=mid] (11) at (1.75, 1.25) {\lang{the truth}};
	\end{pgfonlayer}
	\begin{pgfonlayer}{edgelayer}
		\draw (0.center) to (8.center);
		\draw [bend right=90, looseness=1.25] (4.center) to (1.center);
		\draw [bend right=90, looseness=1.25] (2.center) to (3.center);
	\end{pgfonlayer}
\end{tikzpicture}  
\end{center}

\paragraph{Meaning categories} 
As a first example we describe how pregroup grammar is mapped to $\FVect$, the category of vector spaces and linear transformations.  The noun type $n$ is mapped to a vector space $N$ and the sentence type $s$ to $S$. The concatenation operation in the grammar is mapped to $\otimes$, i.e., the tensor product of vector spaces. Then the morphisms $\epsilon^r_x$ and  $\epsilon^l_x$ map to tensor contraction, and $\eta^r_x$ and $\eta^l_x$ map to identity matrices. 

Function words like verbs and adjectives are modelled as (multi)linear maps. Intransitive verbs are represented as maps from $N$ to $S$, or matrices in $N\otimes S$, and transitive verbs are represented as maps from two copies of $N$ to $S$, or tensors in $N \otimes S \otimes N$. So, in the example above, $\lang{Clowns}$ is mapped to a vector in $N$, as is $\lang{the truth}$, and $\lang{tell} $ is mapped to a tensor in $N \otimes S \otimes N$. The vectors and tensors are concatenated using the tensor product, and tensor contraction is applied to map the sentence down into one sentence vector.

 Compact closed categories have a nice diagrammatic calculus, described in \cite{selinger2010}, or for a linguistically couched explanation see \cite{coecke2010}. In this calculus, the composition of the words $\lang{Clowns}$, $\lang{tell}$, and $\lang{the truth}$ into the sentence $\lang{Clowns tell the truth}$ is expressed as follows:
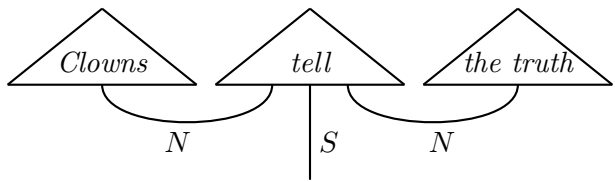
\begin{center}
\begin{tikzpicture}[{every path/.style}=thick]
	\begin{pgfonlayer}{nodelayer}
		\node [style=none, anchor=mid] (0) at (-2.75, 1) {\lang{Clowns}};
		\node [style=none, anchor=mid] (1) at (0, 1) {\lang{tell}};
		\node [style=none, anchor=mid] (2) at (2.75, 1) {\lang{the truth}};
		\node [style=none] (3) at (-2.75, 0.75) {};
		\node [style=none] (4) at (-0.5, 0.75) {};
		\node [style=none] (5) at (0, 0.75) {};
		\node [style=none] (6) at (0.5, 0.75) {};
		\node [style=none] (7) at (2.75, 0.75) {};
		\node [style=none] (8) at (0, -0.5) {};
		\node [style=none] (9) at (4, 0.75) {};
		\node [style=none] (10) at (1.5, 0.75) {};
		\node [style=none] (11) at (2.75, 1.75) {};
		\node [style=none] (12) at (-2.75, 1.75) {};
		\node [style=none] (13) at (-4, 0.75) {};
		\node [style=none] (14) at (-1.5, 0.75) {};
		\node [style=none] (15) at (0, 1.75) {};
		\node [style=none] (16) at (-1.25, 0.75) {};
		\node [style=none] (17) at (1.25, 0.75) {};
		\node [style=none] (18) at (-1.75, 0) {$N$};
		\node [style=none] (19) at (1.75, 0) {$N$};
		\node [style=none] (20) at (0.25, 0) {$S$};
	\end{pgfonlayer}
	\begin{pgfonlayer}{edgelayer}
		\draw [bend right=90, looseness=0.75] (3.center) to (4.center);
		\draw [bend right=90, looseness=0.75] (6.center) to (7.center);
		\draw (5.center) to (8.center);
		\draw [style=swap] (11.center) to (9.center);
		\draw [style=swap] (9.center) to (10.center);
		\draw [style=swap] (10.center) to (11.center);
		\draw [style=swap] (12.center) to (14.center);
		\draw [style=swap] (14.center) to (13.center);
		\draw [style=swap] (13.center) to (12.center);
		\draw [style=swap] (15.center) to (17.center);
		\draw [style=swap] (17.center) to (16.center);
		\draw [style=swap] (16.center) to (15.center);
	\end{pgfonlayer}
\end{tikzpicture}
\end{center}
We will use this notation later to describe how to build particular representations of verbs and other function words.

\subsection{Modelling words as positive operators}
In \cite{piedeleu2015}, \cite{bankova2019}, and \cite{balkir2016} the DisCoCat model is instantiated with the meaning category $\CPM{\FVect}$. This has the same objects as $\FVect$, but the morphisms are now completely positive maps. The $\mathbf{CPM}$ construction is introduced in \cite{selinger}. Words are now represented as positive operators rather than as vectors, and maps between them are completely positive maps. A positive operator is defined as follows, using bra-ket notation from physics.
For a unit~vector $\ket{v}$, the projection operator~$\ket{v}\bra{v}$ onto the subspace spanned by~$\ket{v}$ is called a \emph{pure~state}. A positive operator is given by sum of pure~states. It is an operator $A$ such that:
\begin{enumerate}
\item $\forall{v} \in V.  \braket{v|A|v} \geq 0$,
\item $A$ is self-adjoint
\end{enumerate}
If, in addition, $A$ has trace 1, then $A$ encodes a probabilistic mixture of pure states, and is called a density matrix.
Relaxing this condition gives us different choices for normalization.

Completely positive maps are linear maps that preserve positivity of operators and do so for any trivial extension. 

We give an informal description of how pregroup grammar maps into the category $\CPM{\FVect}$. For more details see \cite{piedeleu2015}, \cite{bankova2019}, or \cite{balkir2016}. Within $\CPM{\FVect}$, the objects are vector spaces, and morphisms are completely positive maps. The underlying spaces that we represent nouns, sentences, and other words in are now doubled up, meaning that a noun is a positive operator $N\rightarrow N$, or a positive semidefinite matrix in $N^* \otimes N$. Morphisms are completely positive maps. These are defined in \cite{selinger} as a morphism $\phi:A^* \otimes A \rightarrow B^* \otimes B$ such that there exists  an object $C$ in the underlying category, in our case $\FVect$, and a morphism $k: C\otimes A \rightarrow B$ such that:

\[
\phi = (k_*\otimes k)\circ (1_{A^*} \otimes \eta_C \otimes 1_A)
\]

Importantly, $\CPM{\FVect}$ is also compact closed, so that the same sort of functorial mapping can be made from the grammar category to the semantics category. Furthermore, the diagrammatic calculus can also be used in this context.



Positive operators were proposed in \cite{balkir2016, bankova2019} as a means of representing word meanings since they have a natural ordering called the L\"owner ordering. This ordering states that for two positive operators $A$ and $B$,
\[
A \loewner B \iff A - B \text{ is positive}
\]
This ordering can be used to represent hyponymy and lexical entailment. In \cite{balkir2016, lewis2019} concrete proposals for building positive operators representing words are given.

The space of positive operators and the properties of the L\"owner ordering on this space has been examined in \cite{dhondt2006, van2016}. When the set of positive operators is restricted to those with maximum eigenvalue less than or equal to 1, the ordering has nice properties. We restrict to this set, and use the notation $\CPe{V}$. When the set of positive operators is restricted to those with eigenvalues exactly 1, we have the projectors, and the L\"owner ordering corresponds to subspace inclusion on projection operators.

The L\"owner ordering is crisp: either the relation obtains or it doesn't. However, when considering natural language, we are also interested in graded notions of hyponymy and entailment. For example, although we may consider $\lang{dog}$ to be highly indicative of $\lang{pet}$, not every dog is a pet, and so we want some kind of graded ordering. On the other hand, we would expect $\lang{dog}$ to be a full hyponym of $\lang{mammal}$

\cite{balkir2016} introduce a graded notion of hyponymy based on the relative entropy of two operators. \cite{bankova2019} use a graded notion of hyponymy that is based on expanding the hypernym (the broader term) to include the hyponym. \cite{lewis2019} extends this idea to include a wider range of gradings.

Specifically, suppose we are comparing two positive operators $A$ and $B$. If $A \loewner B$ crisply, then $B = A + D$ for some positive operator $D$. However, if this is not the case, then we can consider an error term $E$ so that now 
\[
A + D = B + E
\]
Then we have that $B - A = D - E$, i.e. that there is a wholly positive and a wholly negative component to the difference $B - A$. In \cite{bankova2019} the authors render the error term $E$ as being of the form $(1-k)A$, where $k \in [0, 1)$. Then the value $k$ is the strength of the hyponymy relation between $A$ and $B$. The drawback of this approach is that the span of $A$ must be included within the span of $B$. \cite{lewis2019} proposes two alternative gradings based on the error term that do not suffer from this drawback:
\begin{align}
\label{eq:kBA} k_{BA}(A, B) &= \frac{Tr(D - E)}{Tr(D + E)}\\
\label{eq:kE} k_E(A, B) &= 1 - \frac{||E||}{||A||}
\end{align}

In equation \eqref{eq:kBA}, in the worst case the positive difference term $D$ is 0, and then $k_{BA} = -1$. In the best case $E = 0$ and then $k_{BA} = 1$. In equation \eqref{eq:kE}, in the worst case $E = A$, and then $k_E = 0$. In the best case $E = 0$ and then $k_E = 1$.

\subsection{Building positive operators for words}
\label{sec:building}
In \cite{bankova2019}, a broader term such as $\lang{mammal}$ is viewed as a weighted sum over projectors describing instances of mammals. For example:
\begin{align*}
  \semantics{\lang{mammal}} = & p_d\ket{\lang{dog}}\bra{\lang{dog}} + p_c\ket{\lang{cat}}\bra{\lang{cat}} + p_w\ket{\lang{whale}}\bra{\lang{whale}} + ...\\\
  &\mbox{where}\quad \forall i. p_i \geq 0 \text{ (and some kind of normalisation may be applied)}
\end{align*}
\cite{lewis2019} propose a means of building positive operators for words using distributional word vectors and information about hyponymy relations from resources such as WordNet \cite{wordnet}, as follows. In general, the meaning of a word $w$ is considered to be given by a collection of unit vectors~$\{\ket{w_i} \in W\}_i$,
where each~$\ket{w_i}$ represents an instance of the concept expressed by the word.
Then the operator:
\begin{equation}
\label{eq:basic}
\semantics{w} = \sum_i p_i \ \ket{w_i}\bra{w_i} \in W \otimes W
\end{equation}
represents the word~$w$. The $p_i$ are weightings derived from the text, and there are various choices about what these should be.

We build representations of words as positive operators in the following manner. Suppose we have a dictionary of word vectors $\{v_i : \ket{v_i} \in W\}_i$ derived from a corpus using standard distributional or embedding techniques, for example GloVe, \cite{glove}, FastText \cite{fasttext}, or  weighted co-occurrence vectors. To build a representation of a word, we obtain a set of hyponyms that are instances of that word. In this paper, we use WordNet \cite{wordnet}, a human-curated database of word relationships including hyponym-hypernym pairs. The WordNet hyponymy relationship is naturally arranged as a directed graph with a root (it is not quite a tree). For the noun subset of the database, the root is the most general noun \lang{entity}, and the leaves are specific nouns. For example, under the word \lang{rocket} there are (inter alia): \lang{test\_instrument\_vehicle}, \lang{Stinger}, \lang{takeoff\_booster}, \lang{arugula}. Notice that here we have different meanings of the word \lang{rocket}, one as a projectile and one as a vegetable. There are also less supervised ways of obtaining these relationships using patterns derived from text, see \cite{hearst1992, hearst} for examples.

To build a positive operator for a word $w$, we go through the WordNet hierarchy and collect all hyponyms $w_i$ of $w$ at all levels. We then form $\semantics{w}$ as in equation \eqref{eq:basic}, with $p_i = 1$ for all $i$. 

When we build these operators, between 1/3 and 1/2 of the hyponyms listed in WordNet are available in GloVe, and we therefore miss a large proportion of the information included in WordNet.

\subsection{Normalization}
\label{sec:norm}
An important parameter choice is the type of normalization to use. In \cite{bankova2019} two choices are discussed: normalizing operators to trace 1, or normalizing operators to have maximum eigenvalue less than or equal to 1.
The properties of these two normalization strategies are thoroughly analyzed in \cite{vdwetering2017}. If operators are normalized to trace 1, then the crisp L\"owner ordering becomes trivial: no two operators stand in the relation $A \sqsubseteq B$. If operators are normalized to have maximum eigenvalue 1, then the L\"owner ordering has particularly nice properties. In the current paper, we will need to normalize operators so that their maximum eigenvalue is less than or equal to 1, as this will allow us to apply our proposed negation operator.

\subsection{Composing positive operators}
\label{sec:compose}
Building positive operators as proposed gives us representations for individual words. However, the representations are all states in one object of $\CPM{\FVect}$, whereas for verbs, adjectives, and so on, we need morphisms in $\CPM{\FVect}$. In order to obtain these, we use an approach outlined in \cite{Kartsaklis2012}. Firstly, we consider the spaces for noun and sentence to be the same, so now our pregroup types $n$ and $s$ both map to the same space $W$. To represent adjectives and verbs, representations of type $W \otimes W$ or $W \otimes W \otimes W$ are needed. In order to encode our representations in $W \otimes W$ , we need to use the word representations we have built to define suitable morphisms in $\CPM{\FVect}$.
\cite{Kartsaklis2012}  use the notion of a \emph{Frobenius algebra}. Working in $\FVect$, a Frobenius algebra over a finite-dimensional vector space with bases $\{\ket{i}\}_i$ is given by 
\[
\Delta:: \ket{i} \mapsto \ket{i} \otimes \ket{i} \qquad \iota:: \ket{i} \mapsto 1 \qquad \mu:: \ket{i} \otimes \ket{i} \mapsto \ket{i}  \qquad \xi:: 1 \mapsto \ket{i}
\]

In the graphical calculus, these are given by:

\[
\Delta:\begin{gathered}\begin{tikzpicture}[scale=0.5]
	\begin{pgfonlayer}{nodelayer}
		\node [style=none] (0) at (-1, -2.5) {};
		\node [style=none] (1) at (1, -2.5) {};
		\node [style=none] (2) at (-1, -2) {};
		\node [style=none] (3) at (1, -2) {};
		\node [style=small_node] (4) at (0, -1) {};
		\node [style=none] (5) at (0, 0) {};
		\node [style=none] (6) at (0, 0.5) {};
	\end{pgfonlayer}
	\begin{pgfonlayer}{edgelayer}
		\draw [ultra thick, bend right=45, looseness=1.25] (4) to (2.center);
		\draw [ultra thick, bend left=45, looseness=1.25] (4) to (3.center);
		\draw [ultra thick] (5.center) to (4);
	\end{pgfonlayer}
\end{tikzpicture}\end{gathered} \qquad \qquad \iota: \begin{gathered}\begin{tikzpicture}{scale=0.5}
	\begin{pgfonlayer}{nodelayer}
		\node [style=small_node] (0) at (0, -1) {};
		\node [style=none] (1) at (0, 0) {};
		\node [style=none] (2) at (0, 0.25) {};
	\end{pgfonlayer}
	\begin{pgfonlayer}{edgelayer}
		\draw [ultra thick] (1.center) to (0);
	\end{pgfonlayer}
\end{tikzpicture}\end{gathered} \qquad \qquad \mu: \begin{gathered}\begin{tikzpicture}[scale=0.5]
	\begin{pgfonlayer}{nodelayer}
		\node [style=none] (0) at (-1, 0.5) {};
		\node [style=none] (1) at (1, 0.5) {};
		\node [style=none] (2) at (-1, 0) {};
		\node [style=none] (3) at (1, 0) {};
		\node [style=small_node] (4) at (0, -1) {};
		\node [style=none] (5) at (0, -2) {};
		\node [style=none] (6) at (0, -2.5) {};
	\end{pgfonlayer}
	\begin{pgfonlayer}{edgelayer}
		\draw [ultra thick, bend left=45, looseness=1.25] (4) to (2.center);
		\draw [ultra thick, bend right=45, looseness=1.25] (4) to (3.center);
		\draw [ultra thick] (5.center) to (4);
	\end{pgfonlayer}
\end{tikzpicture}\end{gathered} \qquad \qquad \xi:  \begin{gathered}\begin{tikzpicture}
	\begin{pgfonlayer}{nodelayer}
		\node [style=small_node] (0) at (0, 0.25) {};
		\node [style=none] (1) at (0, -0.75) {};
		\node [style=none] (2) at (0, -1) {};
	\end{pgfonlayer}
	\begin{pgfonlayer}{edgelayer}
		\draw [ultra thick] (1.center) to (0);
	\end{pgfonlayer}
\end{tikzpicture}\end{gathered}
\]

A vector $\ket{v} \in W$ can be lifted to a higher-order representation in $ W \otimes W$ by applying the map $\Delta$. 
In $\FVect$, this higher-order representation takes the vector $\ket{v}$ and embeds it along the diagonal of a matrix in $W \otimes W$. So, for example, given a vector representation of an intransitive verb $\ket{\lang{run}} \in W$, we can lift that representation to a matrix in $W \otimes W$ by embedding it into the diagonal of a matrix. 
The Frobenius algebra interacts with the type reduction morphism $\epsilon_N$ in such a way that the result of lifting a verb and then composing with a noun is to apply the $\mu$ multiplication to the tensor product of the noun and the verb vectors, i.e.
\[
(\epsilon_N \otimes 1_N)\circ(1_N \otimes \Delta_N) (\ket{\lang{noun}} \otimes \ket{\lang{verb}}) = \mu (\ket{\lang{noun}} \otimes \ket{\lang{verb}})
\]

Diagrammatically, 
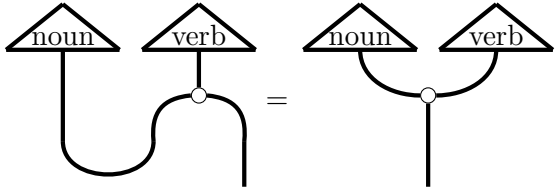
\[
\begin{gathered}\begin{tikzpicture}[scale=0.6]
	\begin{pgfonlayer}{nodelayer}
		\node [style=none] (0) at (1, -3) {};
		\node [style=none] (1) at (-1, -2.5) {};
		\node [style=none] (2) at (1, -2) {};
		\node [style=none] (3) at (-1, -2) {};
		\node [style={small_node}] (4) at (0, -1) {};
		\node [style=none] (5) at (0, 0) {};
		\node [style=none] (6) at (1.25, 0) {};
		\node [style=none] (7) at (-1.25, 0) {};
		\node [style=none] (8) at (0, 1) {};
		\node [style=none, anchor=mid] (9) at (0, 0.25) {verb};
		\node [style=none] (10) at (-4.25, 0) {};
		\node [style=none] (11) at (-1.75, 0) {};
		\node [style=none] (12) at (-3, -2) {};
		\node [style=none] (13) at (-3, 0) {};
		\node [style=none, anchor=mid] (14) at (-3, 0.25) {noun};
		\node [style=none] (15) at (-3, 1) {};
	\end{pgfonlayer}
	\begin{pgfonlayer}{edgelayer}
		\draw [ultra thick, bend left=45, looseness=1.25] (4) to (2.center);
		\draw [ultra thick, bend right=45, looseness=1.25] (4) to (3.center);
		\draw [ultra thick] (5.center) to (4);
		\draw [ultra thick] (8.center) to (6.center);
		\draw [ultra thick] (6.center) to (7.center);
		\draw [ultra thick] (7.center) to (8.center);
		\draw [ultra thick] (13.center) to (12.center);
		\draw [ultra thick] (15.center) to (11.center);
		\draw [ultra thick] (11.center) to (10.center);
		\draw [ultra thick] (10.center) to (15.center);
		\draw [ultra thick, bend left=90, looseness=1.25] (3.center) to (12.center);
		\draw [ultra thick] (2.center) to (0.center);
	\end{pgfonlayer}
\end{tikzpicture}\end{gathered} = \begin{gathered}\begin{tikzpicture}[scale=0.6]
	\begin{pgfonlayer}{nodelayer}
		\node [style=none] (0) at (-1.5, -3) {};
		\node [style={small_node}] (1) at (-1.5, -1) {};
		\node [style=none] (2) at (-3, 0) {};
		\node [style=none] (3) at (-4.25, 0) {};
		\node [style=none] (4) at (-1.75, 0) {};
		\node [style=none] (5) at (-3, 1) {};
		\node [style=none, anchor=mid] (6) at (0, 0.25) {verb};
		\node [style=none] (7) at (1.25, 0) {};
		\node [style=none] (8) at (-1.25, 0) {};
		\node [style=none] (9) at (0, 0) {};
		\node [style=none, anchor=mid] (10) at (-3, 0.25) {noun};
		\node [style=none] (11) at (0, 1) {};
	\end{pgfonlayer}
	\begin{pgfonlayer}{edgelayer}
		\draw [ultra thick, in=180, out=-90, looseness=1.00] (2.center) to (1);
		\draw [ultra thick] (5.center) to (3.center);
		\draw [ultra thick] (3.center) to (4.center);
		\draw [ultra thick] (4.center) to (5.center);
		\draw [ultra thick] (11.center) to (8.center);
		\draw [ultra thick] (8.center) to (7.center);
		\draw [ultra thick] (7.center) to (11.center);
		\draw [ultra thick, in=0, out=-90, looseness=1.00] (9.center) to (1);
		\draw [ultra thick] (1) to (0.center);
	\end{pgfonlayer}
\end{tikzpicture}\end{gathered} 
\]

In $\FVect$ the multiplication $\mu$ implements pointwise multiplication of the two vectors. In $\CPM{\FVect}$ we have access to the same algebra, and the multiplication $\mu$ operates similarly - namely, given two positive operators $A$ and $B$, $\mu(A\otimes B)$ implements pointwise multiplication of the two operators. We call this operator $\textbf{Mult}$ or $\odot$. Whilst simple and theoretically motivated, this operation is not desirable for linguistic purposes as it is commutative, so that `dog bites man' gets the same representation as `man bites dog'.

In \cite{mots, lewisfthc}, two other multiplications are proposed for combining positive operators. One, which we call $\textbf{BMult}$ or $\ast_B$, was originally proposed in \cite{leifer2008, leifer2013} as a quantum Bayesian operation. This takes two operators $A$ and $B$ and returns the non-commutative and non-associative product $B ^\frac{1}{2} A B^\frac{1}{2}$. In \cite{meaningupdating}, the authors show that this operation is also related to a Frobenius algebra, with the caveat that the algebra corresponds to a basis for $W$ that diagonalises $B$.

The second, which we call $\textbf{KMult}$,  or $\ast_K$, is to form a completely positive map from a positive matrix $B$ by decomposing $B$ into a weighted sum of orthogonal projectors $B= \sum_i p_iP_i$, and then forming the map
\[
\mathcal{B}(-) = \sum_i p_i P_i \circ - \circ P_i
\]
If we again consider a basis that diagonalises $B$, this operation then corresponds to the Frobenius multiplication $\mu(A \otimes B)$ in that basis. To see this, consider 
\[
B = \sum_i b_i \ket{i} \bra{i}, \quad A = \sum_{jk} a_{jk} \ket{j} \bra{k}
\]
Then 
\begin{align}
\mathcal{B}(A) &= \sum_i b_i \ket{i} \bra{i} \circ \left(  \sum_{jk} a_{jk} \ket{j} \bra{k} \right) \circ  \ket{i} \bra{i}\\
&= \sum_i b_i \ket{i} \bra{i} \circ   \sum_{j} a_{ji} \ket{j} \bra{i}\\
&= \sum_{i} b_i a_{ii} \ket{i} \bra{i} = \mu(A \otimes B)
\end{align}

We therefore have three ways of combining positive operators. Moreover, each of these combination methods preserves the property that the eigenvalues must be less than or equal to 1. For the operations \textbf{Mult} and \textbf{KMult}, the spectral radius is submultiplicative with respect to the Hadamard (pointwise) product of two positive semidefinite matrices \cite{HornJohnson1985}, implying that the maximum eigenvalue of $A \odot B$ is bounded by 1. For the case of \textbf{BMult}, note that the product $B^\frac{1}{2} A B^\frac{1}{2}$ is similar to $AB$ and hence has the same eigenvalues. Then the maximum eigenvalue of the product $AB$ is bounded by the product of the maximum eigenvalues of $A$ and of $B$ \cite{bhatia2013}, again implying that the maximum eigenvalue of $AB$ is bounded by 1.

To apply these multiplications linguistically, choices must be made about the order in which they are applied, since neither \textbf{BMult} nor \textbf{KMult} are associative. In particular for transitive verbs there are a number of different choices, and some of these are discussed in \cite{lewis2019}. For now, we limit to simple intransitive sentences, of the form $\lang{noun verb}$.

The operators we outlined above are summarised below.
\begin{align}
\label{eq:mult}
\text{Mult: }& \semantics{\lang{noun verb}} = \semantics{\lang{noun}} \odot \semantics{\lang{verb}}\\
\label{eq:bmult}
\text{BMult: }& \semantics{\lang{noun verb}} = \semantics{\lang{noun}}\ast_B\semantics{\lang{verb}}  = \semantics{\lang{verb}}^\frac{1}{2}\semantics{\lang{noun}}\semantics{\lang{verb}}^\frac{1}{2}\\
\label{eq:kmult}
\text{KMult: }& \semantics{\lang{noun verb}} = \semantics{\lang{noun}}\ast_K\semantics{\lang{verb}} = \sum_i p_i P_i \semantics{\lang{noun}} P_i
\end{align}
where in KMult $\semantics{\lang{verb}} = \sum_i p_i P_i$.

\section{Modelling negation in $\CPe{V}$}
So far, we have shown how to build positive operators from a corpus of text, together with information about hyponymy relations. We have also shown how to lift the simple operators thus described to the maps required for functional words such as verbs and adjectives. We now describe how to model negation.

As discussed, one approach to modelling negation is to map a vector to the subspace orthogonal to it. We can incorporate this in our model very easily, since in the case of projectors, this is equivalent to subtracting the associated matrix from the identity matrix. Consider a vector $\ket{dog}$ that we have learnt in a distributional manner from a corpus. We can lift this representation to a positive operator by forming the projector $\ket{dog}\bra{dog}$, which forms a one-dimension subspace of the vector space W. We can then form an operator 
\[
\semantics{\lang{not dog}} = \mathbb{I} -  \ket{dog}\bra{dog}
\]
 which encompasses the $n - 1$-dimensional subspace orthogonal to the projector $\ket{dog}\bra{dog}$. In the general case, we define
\begin{equation}
\semantics{\lang{not w} } := \mathbb{I} - \semantics{\lang{w} }
\end{equation}

When we restrict to the subset $\CPe{W}$ over a vector space $W$, this operation preserves positivity of the operator and also maps operators into the set $\CPe{W}$.

Importantly, this operation is \emph{not} a morphism of $\CPM{\FVect}$, and therefore a suitable home needs to be found for it. We do not provide an answer to that in this paper, leaving it for ongoing work. Rather, we look at how this operation interacts with composition, the L\"owner ordering, and how it works in implementation.

\subsection{How \emph{not} interacts with the (graded) L\"owner ordering}
Consider operators $A$ and $B \in \CPe{W}$. Under the crisp L\"owner ordering,  we have 
\begin{align}
A \loewner B &\iff B = A + D\\
	&\iff \mathbb{I} - B = \mathbb{I} - (A + D)\\
	&\iff \mathbb{I} - B + D= \mathbb{I} - A \iff \lang{not }B \loewner \lang{not }A
\end{align}
Considering an error term $E$, we use the notation $\loewner_E$ if $B + E = A + D$. With such an error term,
\begin{align}
A \loewner_E B &\iff B + E = A + D\\
	&\iff \mathbb{I} - (B + E) = \mathbb{I} - (A + D)\\
	&\iff \mathbb{I} - B + D= \mathbb{I} - A + E\iff \lang{not }B \loewner_E \lang{not }A
\end{align}
Depending on the grading we use, the strength of the hyponymy relation will be affected. Using the $k_{BA}$ grading (equation \eqref{eq:kBA}) we have that 
$\lang{not } B$ is a hyponym of $\lang{not } A$ with strength 
\[
k_{BA} (\lang{not } B, \lang{not } A) = \frac{Tr(D - E)}{Tr(D + E)} = k_{BA}(A, B)
\]
Using $k_E$ (equation \eqref{eq:kE}), we have:
\[
k_{E} (\lang{not } B, \lang{not } A) = 1 - \frac{||E||}{||\lang{not } B||} \neq k_{E}(A, B)
\]

\subsection{How \emph{not} interacts with composition}
We focus here just on the case of intransitive sentences composed of a subject and a verb. When we negate the noun we obtain the following expressions:
\begin{align}
\semantics{\lang{not noun}} \odot \semantics{verb} &= (\I - \semantics{\lang{noun}})\odot \semantics{verb}\\
&= diag(\semantics{\lang{verb}}) - \semantics{\lang{noun}} \odot \semantics{verb}
\end{align}
\begin{align}
\semantics{\lang{not noun}} \ast_B \semantics{verb} &= (\I - \semantics{\lang{noun}})\ast_B \semantics{verb}\\
&= \semantics{\lang{verb}}^\frac{1}{2}\semantics{\lang{verb}}^\frac{1}{2} - \semantics{\lang{verb}}^\frac{1}{2}\semantics{\lang{noun}}\semantics{\lang{verb}}^\frac{1}{2}\\
&= \semantics{\lang{verb}} - \semantics{\lang{noun}} \ast_B \semantics{verb}
\end{align}
\begin{align}
\semantics{\lang{not noun}} \ast_K \semantics{verb} &= (\I - \semantics{\lang{noun}})\ast_K \semantics{verb}\\
&= \sum_i p_i P_i P_i - \sum_i p_i P_i \semantics{\lang{noun}} P_i\\
&= \semantics{\lang{verb}} - \semantics{\lang{noun}} \ast_K \semantics{verb}
\end{align}
Particularly in the case of $\ast_B$ and $\ast_K$, these feel like fairly natural interpretations of a sentence with a negated noun. We take the meaning of the verb as a whole, and then subtract out the part of the verb that is applied to the noun.

When we negate the verb we obtain the following expressions:
\begin{align}
\semantics{\lang{noun}} \odot \semantics{\lang{not verb}} &= \semantics{\lang{noun}}\odot (\I - \semantics{verb})\\
&= diag(\semantics{\lang{noun}}) - \semantics{\lang{noun}} \odot \semantics{verb}
\end{align}
and, assuming that we use a basis in which $\semantics{\lang{verb}}$ is diagonal:
\begin{align}
\semantics{\lang{noun}} \ast_K \semantics{\lang{not verb}} &= (\semantics{\lang{noun}})\ast_K (\I - \semantics{verb})\\
&= \sum_i (1 - p_i) P_i \semantics{\lang{noun}} P_i\\
&= \sum_i P_i \semantics{\lang{noun}} P_i\  - \sum_i p_i P_i \semantics{\lang{noun}} P_i\\
&=diag (\semantics{\lang{noun}}) - \semantics{\lang{noun}} \ast_K \semantics{verb}
\end{align}

The operation $\ast_B$ does not have a particularly illuminating representation when the verb is negated, but in the case of $\odot$ and $\ast_K$, these are again fairly natural interpretations of a sentence with a negated verb.

\section{Demonstrations}
We give a demonstration on a small dataset that this rendering of negation works well together with the composition operators proposed. In particular, we will see that our combination operators can beat baselines that examine just the noun or the verb in the sentence. This is an important baseline since the construction of the dataset is such that entailment does follow from comparing either the nouns or the verbs. Our combination operators do not in general beat an average of two operators, however, they do in some cases.

\subsection{Datasets}
We build a set of datasets based on the intransitive sentence dataset introduced in \cite{balkir2015b}. The dataset consists of paired sentences consisting of a subject and a verb. In half the cases the first sentence entails the second, and in the other half of cases, the order of the sentences is reversed. For example, we have:
\begin{quote}
summer finish, season end, T

season end, summer finish, F
\end{quote}
The first sentence is marked as entailing, whereas the second is marked as not entailing. The dataset is created by selecting nouns and verbs from WordNet. In the case of the sentence marked T, the first noun is selected as a hyponym of the second noun, and the first verb is selected as a hyponym of the second verb. 

For these sentences to be thought of as entailing, we must view them as being implicitly existentially quantified. For example, if we took the pair of sentences 
\begin{quote}
gazelles sprint, mammals run
\end{quote}
we can clearly see that the first sentence does not entail the second if we assume a universal quantification - there could easily be, and there are, non-gazelle mammals that don't run. However, if we take an existential quantification, then the fact that there is some gazelle that sprints means that there must be some mammal (the gazelle) who runs (as  sprinting is a kind of running).

Bearing in mind that the sentences are existentially quantified, we create three further datasets that include negation. We apply negation only at the word level and not at the sentence level, as this retains the existentially quantified nature of the sentences. Consider an entailing sentence pair such as:
\[
\lang{dogs run} \models \lang{mammals move}
\]
We include negation in two places: either the noun can be negated, giving us \lang{non-dogs} and \lang{non-mammals}, or else the verbs can be negated, giving us \lang{do not run} and \lang{do not move}. 

From $\lang{dogs run} \models \lang{mammals move}$ we then get three more pairs of entailing sentences:
\begin{align}
\lang{some dogs run} &\models \lang{some mammals move}\\
\lang{some non-mammals run} &\models \lang{some non-dogs move}\\
\lang{some dogs do not move} &\models \lang{some mammals do not run}\\
\lang{some non-mammals do not move} &\models \lang{some non-dogs do not run}
\end{align}
To model these, we render the negation of the verb as directly acting on the verb. Another choice would be for the negation to act on the whole sentence, rendering \lang{dogs don't move} as \lang{not(dogs move)}, but this would mean that we now consider the sentence universally quantified. Working out how to include  a full account of  quantification is an area of further work. 

To model these sentences, we therefore calculate, respectively:
\begin{align}
\semantics{\lang{dogs}}\ast  \semantics{\lang{run}}&\loewner_k \semantics{\lang{mammals}}\ast  \semantics{\lang{move}}\\
(\I - \semantics{\lang{mammals}})\ast  \semantics{\lang{run}}&\loewner_k (\I - \semantics{\lang{dogs}})\ast  \semantics{\lang{move}}\\
\semantics{\lang{dogs}}\ast  (\I - \semantics{\lang{move}})&\loewner_k \semantics{\lang{mammals}}\ast  (\I - \semantics{\lang{run}})\\
(\I - \semantics{\lang{mammals}})\ast  (\I - \semantics{\lang{move}})&\loewner_k (\I - \semantics{\lang{dogs}})\ast  (\I - \semantics{\lang{run}})
\end{align}
where $\loewner_k \in \{k_{BA}, k_E\}$ is one of the graded hyponymy measures and $\ast \in \{\odot, \ast_B, \ast_K\}$ is one of the compositional operators.

\subsection{Construction and composition of positive operators}
We follow the construction methods outlined in \cite{lewis2019} and summarised in this paper in section \ref{sec:building}. In order to construct the basic positive operators, we use hyponyms from WordNet \cite{wordnet}, and 50 or 300 dimensional GloVe vectors. The operators produced are normalised to have maximum eigenvalue equal to 1.

To compose positive operators, we use the three composition functions \textbf{Mult},  \textbf{Mult},  \textbf{Mult} discussed in section \ref{sec:compose}. We compare these with three baselines: the average of two operators, a noun-only baseline, and a verb-only baseline. Due to the construction of the datasets, we see that in fact the verb-only and noun-only baselines are fairly strong, since as long as the construction of the individual words models the hyponymy relations well then a verb-only or noun-only model will be able to perform well on these datasets. Note that taking the average of the two operators preserves the criterion of the maximum eigenvalue being less than or equal to 1 by Weyl's inequalities \cite{weyl1912}

\paragraph{Metrics and significance measures}
Since the entailment measures we use give back a grading, whereas we require a binary response, we calculate area under ROC curve (AUC). The AUC calculates the true positive rate vs. the false positive rate for different cutoff levels of the graded measure. The maximum that can be attained is 1.

To measure the significance of our results, we use bootstrapping \cite{efron1992} to calculate 100 values of the test statistic (AUC) drawn from the distribution implied by the data. We compare between models using a paired t-test and apply the Bonferroni correction to compensate for multiple model comparisons.

\section{Results}

\begin{table}
\caption{Area under ROC curve on the negation datasets, using $k_{BA}$, WordNet hyponyms, and 300 dimensional GloVe vectors. Figures reported are the average of the 100 values of the test statistic. $^*$ indicates significantly better than the Average baselline. $^+$ indicates significantly better than the noun-only baseline.}
\centering
\small 
\label{tab:ks2016300ba}
\begin{tabular}{l c c c c}
Model & noun-verb & $\neg$noun-verb &   noun-$\neg$verb & $\neg$noun-$\neg$verb\\
\hline
KS2016 best & 0.84 & - & - & - \\
\hline
Verb only    & 0.866	 	& 0.867 	& 0.865 	& 0.867\\
Noun only   & 0.926		& 0.921	&0.925 	&0.923\\
Average      & $0.947^+$ & $\textbf{0.946}^+$ & $\textbf{0.948}^+$ &$0.946^+$\\
\hline
Mult & $\textbf{0.960}^{*+} $& 0.874	 &$0.931^+$ & $\textbf{0.950}^+$\\
BMult & $	     0.948^+ $	  & 0.892	& 0.928	&$0.947^+$\\
BMult switched & $0.949^+$ & 0.896	& 0.916	&$0.944^+$\\
KMult &            $ 0.950^+ $& 0.875	& 0.925 	& $0.948^+$\\
KMult switched &$ 0.950^+$ & 0.874	& 0.920 	& $0.948^+$
\end{tabular}
\end{table}

\begin{table}
\caption{Area under ROC curve on the negation datasets, using $k_{E}$, WordNet hyponyms, and 300 dimensional GloVe vectors. Figures reported are the average of the 100 values of the test statistic. $^*$ indicates significantly better than the Average baselline. $^+$ indicates significantly better than the noun-only baseline.}
\centering
\small 
\label{tab:ks2016300e}
\begin{tabular}{l c c c c}
Model & noun-verb & $\neg$noun-verb &   noun-$\neg$verb & $\neg$noun-$\neg$verb\\
\hline
KS2016 best & 0.84 & - & - & - \\
\hline
Verb only    & 0.635	 	& 0.637 	& 0.636 	& 0.634\\
Noun only   & 0.686		& 0.643	&0.684 	&0.635\\
Average      & $0.727$ & ${0.778}^+$ & ${0.777}^+$ &$0.782^+$\\
\hline
Mult & $\textbf{0.883}^{*+} $& $\textbf{0.885}^{*+}$		&$0.899^{*+}$ & $\textbf{0.952}^{*+}$\\
BMult & $	     0.792^{*+} $	  & $0.678	^+$		& $0.725^+$			&$0.719^+$\\
BMult switched & $0.786^{*+}$ & $0.693^+$	& $0.715^+$			&$0.718^+$\\
KMult &            $ 0.873^{*+} $& $0.725^+$	& $\textbf{0.900}^{*+}$ 	& $0.732^+$\\
KMult switched &$ 0.839^{*+}$ & $0.879^{*+}$	& $0.732^+$ 			& $0.666^+$
\end{tabular}
\end{table}

\begin{table}
\caption{Area under ROC curve on the negation datasets, using $k_{BA}$, WordNet hyponyms, and 50 dimensional GloVe vectors. Figures reported are the average of the 100 values of the test statistic. $^*$ indicates significantly better than the Average baselline. $^+$ indicates significantly better than the noun-only baseline.}
\centering
\small 
\label{tab:ks201650ba}
\begin{tabular}{l c c c c}
Model & noun-verb & $\neg$noun-verb &   noun-$\neg$verb & $\neg$noun-$\neg$verb\\
\hline
KS2016 best & 0.84 & - & - & - \\
\hline
Verb only    & 0.787	 	& 0.787 	& 0.786 	& 0.785\\
Noun only   & 0.907		& 0.906 	&0.903 	&0.904\\
Average      & $0.929^+$ & $\textbf{0.925}^+$ & $\textbf{0.929}^+$ &$\textbf{0.930}^+$\\
\hline
Mult & $\textbf{0.942}^{*+} $& 0.836	 &$0.915^+$ & ${0.925}^+$\\
BMult & $	     0.917^+ $	  & 0.861	& $0.914^+$	&$0.920^+$\\
BMult switched & $0.918^+$ & 0.859	& $0.912^+$	&$0.922^+$\\
KMult &            $ 0.929^+ $& 0.829	& $0.910^+$ 	& $0.926^+$\\
KMult switched &$ 0.926^+$ & 0.821	& $0.911^+$ 	& $0.930^+$
\end{tabular}
\end{table}

\begin{table}
\caption{Area under ROC curve on the negation datasets, using $k_{E}$, WordNet hyponyms, and 50 dimensional GloVe vectors. Figures reported are the average of the 100 values of the test statistic. $^*$ indicates significantly better than the Average baseline. $^+$ indicates significantly better than the noun-only baseline.}
\centering
\small 
\label{tab:ks201650e}
\begin{tabular}{l c c c c}
Model & noun-verb & $\neg$noun-verb &   noun-$\neg$verb & $\neg$noun-$\neg$verb\\
\hline
KS2016 best & 0.84 & - & - & - \\
\hline
Verb only    & 0.601	 	& 0.605 	& 0.605 	& 0.607\\
Noun only   & 0.708		& 0.724	&0.706 	&0.720\\
Average      & $0.753^+$ & ${0.791}^+$ & ${0.783}^+$ &$0.797^+$\\
\hline
Mult & $0.847^{*+} $& $\textbf{0.845}^{*+}$	&$\textbf{0.891}^{*+}$ & $\textbf{0.925}^{*+}$\\
BMult & $	     0.751^{+} $	  & $0.694	$		& $0.738^+$			&$0.751^+$\\
BMult switched & $0.728^{+}$ & $0.707$		& $0.727^+$			&$0.758^+$\\
KMult &  $ \textbf{0.875}^{*+} $& $0.702$		& ${0.875}^{*+}$ 		& $0.802^+$\\
KMult switched &$ 0.808^{*+}$ & $0.791^{*+}$	& $0.726^+$ 			& $0.815^{*+}$
\end{tabular}
\end{table}

We can see that across the board (tables \ref{tab:ks2016300ba}, \ref{tab:ks2016300e}, \ref{tab:ks201650ba}, \ref{tab:ks201650e}), the $k_{BA}$ measure performs more strongly than the $k_E$ measure. The difference in performance is likely to be because the $k_{BA}$ measure is very symmetric, and the dataset is also, meaning that not only are there equal numbers of entailing and non-entailing sentences in the dataset, but the non-entailing datasets are the opposite of the entailing datasets. Enhancing the datasets with some random pairings would likely degrade the performance of the $k_{BA}$ measure. Investigating the differences in performance in a less balanced dataset is an area of further work.

In the case of the $k_{BA}$ measure, increasing the dimensionality of the underlying vector space improved performance across all sentence types. This was not the case for the $k_E$ measure, where for sentences of the type $\lang{noun - not verb}$ and $\lang{not noun - verb}$ performance using the $k_E$ measure improved with lower dimensionality (tables \ref{tab:ks2016300e} and \ref{tab:ks201650e})

The best results were obtained using the $k_{BA}$ measure and 300-dimensional GloVe vectors. In this set of results (table \ref{tab:ks2016300ba}) the Average baseline proves hard to beat, however Mult also performs strongly for sentences with either no word negated or both words negated. For these two classes of sentences, it is also notable that all composition functions enable better performance than the strong non-compositional noun-only baseline. A similar pattern is seen when using 50-dimensional vectors with the $k_BA$ measure (table \ref{tab:ks201650ba}), where the benefit of using a compositional operator is also seen for the sentence type $\lang{noun - not verb}$. 

The benefit of using compositional operators is also seen for the $k_E$ measure (tables \ref{tab:ks2016300e} and \ref{tab:ks201650e}), where using a compositional operator helps in almost all cases over the (admittedly much worse) non-compositional noun-only baseline.

Across both measures and dimensionalities performance is poor on the sentence type $\lang{not noun - verb}$. More research is needed to investigate why this is.



\section{Discussion and Conclusions}
We have introduced a negation operator for use in the $\CPM{\FVect}$ flavour of DisCoCat. The operators is based on the notion of projection onto the orthogonal subspace, used previously by \cite{widdows2003}. The operator works well together with the composition operators Mult, BMult, and KMult discussed in \cite{lewisfthc, lewis2019, meaningupdating}, and in many cases perform well on a toy dataset of sentence entailments.

More investigation into the properties of the \textbf{BMult} and \textbf{KMult} operators is needed. \cite{meaningupdating} have shown that the two operators can be combined together in a double density matrix setting, meaning that the operators can be given a natural home.

Work is also ongoing to build operators from corpora in a less supervised way. Recent work on learning Gaussian embeddings \cite{vilnis2014} may be leveraged to build the representations needed.

Further, testing on larger scale datasets is also needed. Ideally, the kinds of entailment relations we are looking at should be useful for textual entailment and reasoning systems. Expanding the models we currently have to test on realistic datasets is desirable.

Another major unanswered question is where the negation operator should sit theoretically. It cannot be viewed as a morphism in $\CPM{\FVect}$. Some work in progress is into looking at the set $\CPe{W}$ as an object of the category $\ConvexRel$, introduced in \cite{bolt2017}. Then, the negation operator can be viewed as a morphism. This is an area of further work.

\bibliographystyle{plainnat}
\bibliography{hyp}

\end{document}